\def\year{2022}\relax
\documentclass[letterpaper]{article} 
\usepackage{aaai22}  
\usepackage{times}  
\usepackage{helvet}  
\usepackage{courier}  
\usepackage[hyphens]{url}  
\usepackage{graphicx} 
\usepackage{natbib}  
\usepackage{caption} 
\DeclareCaptionStyle{ruled}{labelfont=normalfont,labelsep=colon,strut=off} 
\usepackage{algorithm}
\usepackage{algorithmic}
\usepackage{comment}
\usepackage{multicol}
\usepackage{multirow}
\usepackage{booktabs}
\usepackage{xcolor}
\usepackage{enumitem}
\usepackage{rotating}
\usepackage{graphicx}
\usepackage{pdflscape}
\usepackage{caption}
\usepackage{subcaption}

\usepackage{newfloat}
\usepackage{listings}
\floatstyle{ruled}
\newfloat{listing}{tb}{lst}{}
\floatname{listing}{Listing}
\pdfoutput=1

\title{Is it all a cluster game? - Exploring Out-of-Distribution Detection based on Clustering in the Embedding Space}
\author {
    Poulami Sinhamahapatra,\textsuperscript{\rm 1}
    Rajat Koner, \textsuperscript{\rm 2}
    Karsten Roscher,\textsuperscript{\rm 1}
    Stephan Günnemann \textsuperscript{\rm 3}
}
\affiliations {
    \textsuperscript{\rm 1}Fraunhofer-Institut für Kognitive Systeme IKS,
    \textsuperscript{\rm 2} Ludwig Maximilian University of Munich\\
    \textsuperscript{\rm 3} Technical University of Munich\\
}

\begin{document}

\maketitle

\begin{abstract}
It is essential for safety-critical applications of deep neural networks to determine when new inputs are significantly different from the training distribution. In this paper, we explore this out-of-distribution (OOD) detection problem for image classification  using clusters of semantically similar embeddings of the training data and exploit the differences in distance relationships to these clusters between in- and out-of-distribution data. We study the structure and separation of clusters in the embedding space and find that the supervised contrastive learning leads to well separated clusters while its self-supervised counterpart fails to do so. In our extensive analysis of different training methods, clustering strategies, distance metrics and thresholding approaches, we observe that there is no clear winner. The optimal approach depends on the model architecture and selected datasets for in- and out-of-distribution. While we could reproduce the outstanding results for contrastive training on CIFAR-10 as in-distribution data, we find standard cross-entropy paired with cosine similarity outperforms all contrastive training methods when training on CIFAR-100 instead. Cross-entropy  provides competitive results as compared to expensive contrastive training methods.
\end{abstract}

\section{Introduction}
\label{sec:intro}
The recent success of Deep Neural Networks (DNN) has motivated their application in an varity of tasks. While DNNs have demonstrated remarkable performance, they cannot be expected to work reliably on inputs that are not represented by the training distribution. Such out-of-distribution (OOD) samples can lead to  unpredictable behaviour and overconfident predictions \citep{Nguyen_2015_CVPR, guoCalibrationModernNeural2017a, hendrycksBaselineDetectingMisclassified2018a}, with severe consequences in case of safety-critical applications like autonomous driving or automated medical diagnoses. Therefore, it is crucial to detect such inputs when applied to the model to allow for additional fallback measures to be triggered \citep{henneManagingUncertaintyAIbased2019} or to abstain from automated decisions in rare or unseen situations  \cite{zhouReviewDeepLearning2021, prabhuPrototypicalClusteringNetworks2018}. 

\begin{figure}
    \centering
    \includegraphics[height=3cm]{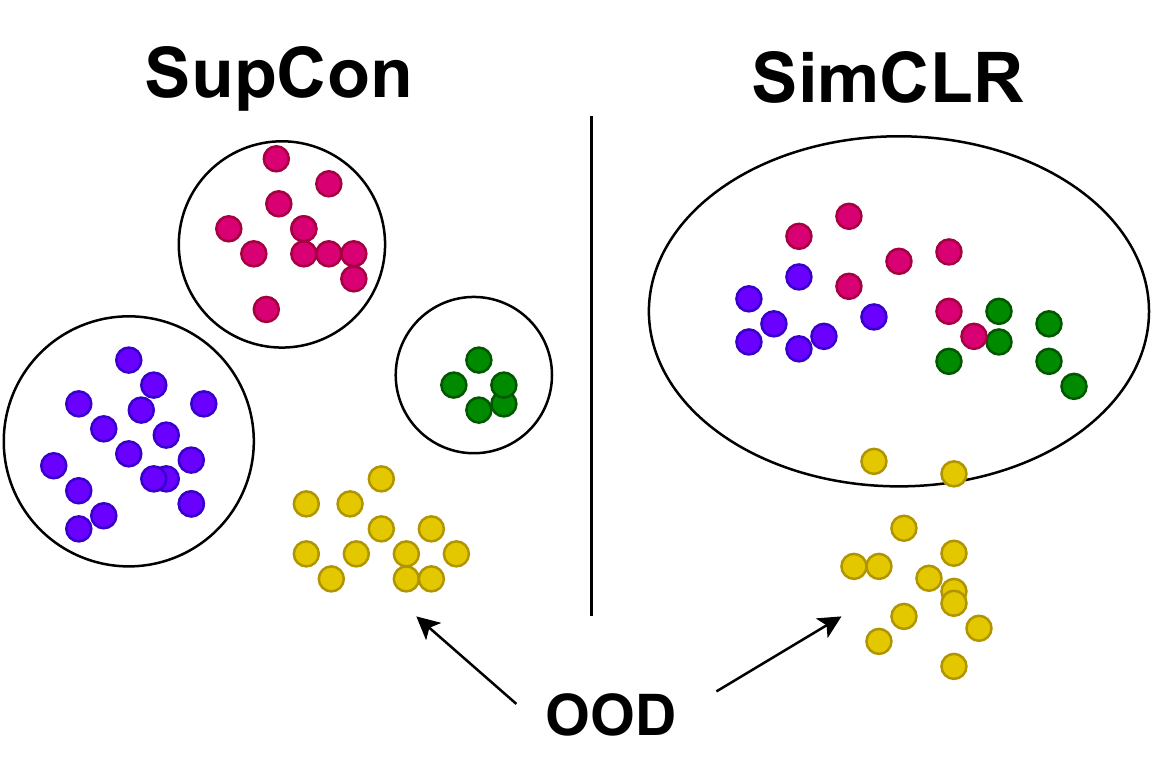}
    \caption{Simplified illustration of the expected latent space clusters for supervised (SupCon) and unsupervised (SimCLR) contrastive training methods
    }
    \label{fig:supcon_simclr}
\end{figure}

One promising research direction for out-of-distribution detection - especially in image classification - is to exploit the distribution of training samples in the learnt embedding space assuming that related images exhibit similar features and are therefore in close proximity according to their latent representation~\cite{leeSimpleUnifiedFramework2018}. Since contrastive learning (CL) methods \cite{khoslaSupervisedContrastiveLearning2020, chenSimpleFrameworkContrastive2020c} are supposed to improve the separability of instances or samples in the embedding space by pulling similar inputs together and pushing dissimilar ones apart, it is only natural that their use for OOD detection based on latent representations has demonstrated state-of-the-art results recently~\cite{sehwagSSDUnifiedFramework2021}. Figure~\ref{fig:supcon_simclr} illustrates the intuition behind those approaches.  

However, while the results are promising, several aspects are left unexplored. On the one hand, there is the question to which extent different training methods really influence the formation of well-defined clusters of in-distribution (ID) samples in the embedding space where different distance metrics may be applied to measure sample similarity. On the other hand, clustering strategies and the optimal number of clusters have barely been touched in existing literature, with the odd choice of a single cluster representing all the ID data apparently leading to the best results~\cite{sehwagSSDUnifiedFramework2021}. Since the use of machine learning in safety-critical contexts depends on a sound understanding of the insufficiencies and expected failure modes of the deployed models \cite{burtonSafetyAssuranceMachine2021}, we conduct an extensive study on the performance of supervised and self-supervised contrastive learning methods for OOD detection focussing on the following contributions:

\begin{itemize}
    \item \textbf{Structure of the embedding space:} In Section \ref{sec:cluster}, we provide detailed insights into cluster formations across supervised (SupCon) and unsupervised (SimCLR) contrastive learning methods, by  using \textit{Global Separation} and \textit{Cluster Purity} metrics to analyse cluster quality. We find that supervised training leads to well-separated clusters, while unsupervised training leads to closely overlapping clusters.
    \item \textbf{OOD detection based on clustering:} In Section \ref{sec:ood}, we provide a modular OOD detection approach based on the similarity of an input sample to a set of clusters allowing the comparison of different distance metrics, clustering methods and thresholding strategies. We further investigate, whether observations are consistent for different models and data sets. Our results indicate that there is no clear winner: the optimal combination indeed depends on the model size, distance metrics and training data. 
\end{itemize}

\section{Related Work} 
\label{sec:recent_work}
DNNs are increasingly used in tasks like classification\cite{dosovitskiy2020image},scene prediction\cite{koner2021scenes,koner2020relation} and other high level tasks such as reasoning \cite{hildebrandt2020scene,koner2021graphhopper}. However, presence of OOD samples presents an important concern in the successful completion of all such tasks, particularly in safety-critical systems. Thus, reliable OOD detection has become an important direction of research.
\paragraph{Out-of-Distribution Detection}
The problem of OOD detection has often been formulated as outlier detection \cite{hodgeSurveyOutlierDetection2004, sehwagSSDUnifiedFramework2021}, one-class classification \cite{ruffDeepOneClassClassification2018, pereraOCGANOneClassNovelty2019}, novelty detection \cite{tackCSINoveltyDetection2020, pidhorskyiGenerativeProbabilisticNovelty2018}, anomaly detection \cite{golanDeepAnomalyDetection2018, hendrycksDeepAnomalyDetection2019} and  open set recognition \cite{boultLearningUnknownSurveying2019, gengRecentAdvancesOpen2020}. 
Some contemporary ways to approach the problem are: 
density approximation based generative modelling \cite{ren2019likelihood, nalisnickDeepGenerativeModels2019}, self-supervision to learn discriminatory features  \cite{hendrycksUsingSelfSupervisedLearning2019, mohseniSelfSupervisedLearningGeneralizable2020, tackCSINoveltyDetection2020, sehwagSSDUnifiedFramework2021},  softmax score based classifier methods \cite{hendrycksBaselineDetectingMisclassified2018a, liangEnhancingReliabilityOutofdistribution2020}, detection score based methods \cite{leeSimpleUnifiedFramework2018, winkensContrastiveTrainingImproved2020d, tackCSINoveltyDetection2020}, utilisation of  uncertainty quantifications based methods \cite{schwaigerUncertaintyQuantificationDeep2020, charpentierPosteriorNetworkUncertainty2020b} as well as methods using self-attention based transformers \cite{koner_oodformer_2021}.
Since OOD samples can vary in many different ways, many outlier exposure methods use few known OOD samples, thus inducing a form of prior knowledge of OOD \cite{ leeSimpleUnifiedFramework2018, hendrycksDeepAnomalyDetection2019, liangEnhancingReliabilityOutofdistribution2020}. 
However, this approach  could lead to problems when generalising across diverse novel OOD datasets. 
 Many contemporary works have explored multi-class OOD detection settings 
 without inducing prior bias for OOD samples, but often do not perform well with only near-OOD data, i.e. semantically similar from ID data. In such a scenario, instance based discriminatory method like CL can be used to learn useful semantic features. 
 

\paragraph{Contrastive Learning:} 
Discriminative approaches using contrastive loss \cite{bachmanLearningRepresentationsMaximizing2019, hjelmLearningDeepRepresentations2019} had shown great promise in the past, however recently CL has found even greater application in multiple application domains \cite{henaffDataEfficientImageRecognition2020, tackCSINoveltyDetection2020, sehwagSSDUnifiedFramework2021} following the success of self-supervised methods like SimCLR \cite{chenSimpleFrameworkContrastive2020c}, MoCov2 \cite{chenImprovedBaselinesMomentum2020a}, etc
as well as Supervised CL method \cite{khoslaSupervisedContrastiveLearning2020} and similar.
Recent works like \cite{sehwagSSDUnifiedFramework2021, tackCSINoveltyDetection2020, winkensContrastiveTrainingImproved2020d} have employed contrastive training for OOD detection by either modifying the contrastive training objective or assuming inherent class-conditioned clusters. 
As a novel contribution, 
we present an extensive study into the quality of clusters formed by various contrastive training approaches and their influence on OOD detection.

\section{Structure of the embedding space} 
\label{sec:cluster}

In this section, we investigate potential clusters in the embedding space and address the question about how to evaluate the quality of clusters as well as their separation in the high-dimensional embedding space.

\subsection{Contrastive Learning towards clustering} \label{sec:cluster_methods}

The key intuition behind any CL method is to preserve a meaningful representation by maximising the agreement between similar instances and at the same time minimising the agreement with dissimilar instances. This means that, given an anchor image and a set of positives and negatives, the positives are pulled closer based on similarity with  the anchors while the negatives are pushed apart in the embedding space. In this work, we focus on Supervised Contrastive Learning (SupCon) \cite{khoslaSupervisedContrastiveLearning2020} and the unsupervised approach SimCLR \cite{chenSimpleFrameworkContrastive2020c} and compare them to a baseline trained with standard cross-entropy (CE) loss.

\textit{SimCLR} uses strong data augmentation to compare with positive instances of an anchor image to learn without supervision. While this promotes discriminative feature learning, it often leads to disagreement between instances of same classes. \textit{SupCon} tries to address this caveat, by increasing the number of positives by using all the samples of the same class/ same ground-truth (GT) labels for comparison with each anchor during CL. Both  methods apply the contrastive loss,  based on cosine similarity,  at a lower-dimensional non-linear projection layer. Since this layer is trained to be invariant to augmentations, the authors \cite{chenSimpleFrameworkContrastive2020c} suggest that good quality representations are most likely to be preserved at the last \textit{feature} layer of the encoder.

\subsection{Determining Cluster Quality }
\label{sec:cluster_quality}

In our investigation of cluster quality, we want to determine how well clusters are separated  from each other. Clusters are either formed based on class labels corresponding to GT classes or by using additional clustering like k-means on the embedding vectors of all training samples. In the case of k-means clustering,  we are further interested to understand if these clusters reflect semantically similar samples, e.g. with samples belonging to the same class.

  \begin{figure*}[h]
    \centering
    \includegraphics[width=0.9\linewidth, height=4cm]{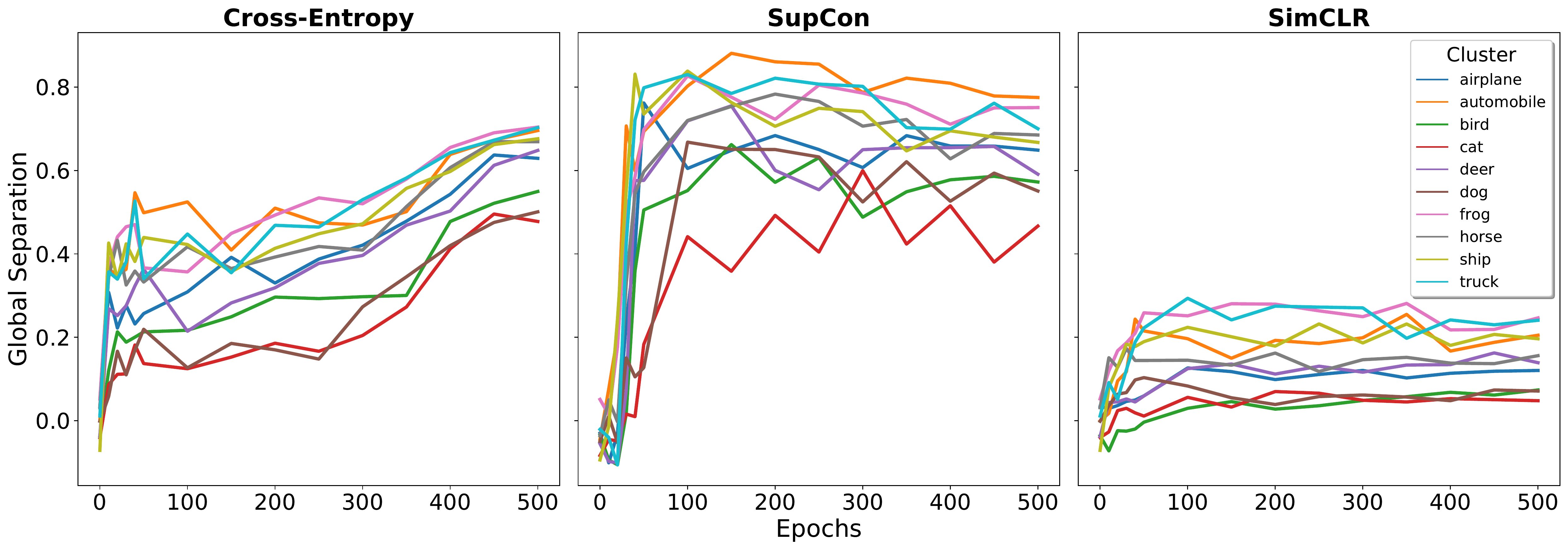}

    \caption{Evolution of Global Separation between clusters based on GT classes over training epochs for different contrastively trained models on CIFAR-10 as compared to baseline CE.}
    \label{fig:evolution}
\end{figure*}

\noindent \textbf{Global Separation (GS)} \cite{bojchevskiRobustSpectralClustering2017} generalises and extends the idea of the Silhouette Coefficient \cite{rousseeuwSilhouettesGraphicalAid1987}. GS utilises the intuition that separability between clusters can be determined by inspecting intra-cluster and inter-cluster distances. Thus, for each cluster $c$, a list of pairwise distances $P_{c,c}$ is calculated for all samples within the same cluster (intra-cluster distance). In addition,  a list of pairwise distances for samples from another cluster $c'$, $P_{c,c'}$ is computed.

Finally, GS for a given cluster, taking smallest $x\%$ samples, is given as the difference between the intra-cluster distances and the distance to the closest different cluster, normalised by maximum of the two values, given as:
\begin{equation}\label{eq:gs}
    \textit{GS}_c (x) = \frac{P_{c,c^{'}} (x) - P_{c,c}(x)}{\textit{max}(P_{c,c^{'}} (x), P_{c,c} (x))}
\end{equation}

\noindent\textbf{Cluster Purity} is used to determine how many samples in a cluster belong to the same class when k-means clustering is applied: 
\begin{equation} \label{eq:cp}
    \textit{CP}_c  = \frac{\textit{max}_j |K_c \cap t_j | } {N_c}
\end{equation}
where $K_c$ , $N_c$ denote the samples in a given k-means cluster and their total count respectively  whereas $t_j$ refers to samples from $j^{th}$ GT class. Say, we assume a k-means cluster with 500 samples that has 490 samples belonging to the same GT class. That would lead to a cluster purity of 98\%.

\subsection{Experiments and Discussion} 
\label{sec:cluster_exp}

Here, we discuss several experiments conducted on the CIFAR-10 dataset \cite{krizhevsky2009learning} using ResNet-50 model \cite{he2016deep}  to evaluate cluster quality of contrastively trained SupCon, SimCLR with CE as baseline using the metrics presented in Section \ref{sec:cluster_quality}.

\paragraph{Evolution of cluster formation over training time:}
 We investigated the evolution of separation of class-based clusters over training epochs as shown in Figure \ref{fig:evolution}. We observe that supervised methods like CE and SupCon show increasing global separation over time starting initially from a negative separation (epoch 0) with SupCon learning much faster and showing better separation. Notably, CE still leads to quite well-defined clusters even without a contrastive loss. The unsupervised SimCLR does not show much further separation after an initial clustering of embeddings, based on GT classes. The unsupervised goal of discriminating between individual samples rather than class, hurts the overall formation of class-based clusters placing it even further below cross-entropy in that domain. 
 Some classes like \lq frog \rq,  \lq automobile \rq or \lq truck \rq show good separability while \lq cat \rq and \lq dog \rq have consistently worse clusters across all methods. This could indicate that \lq cat \rq and \lq dog \rq are harder to distinguish, sharing the majority of overall features, than the other more separated classes. On the other hand, it could also be an artefact of the individual distribution of samples for each class in the CIFAR-10 dataset. 
 
 \begin{figure}[h!]
\centering
    \includegraphics[width=0.45\linewidth, keepaspectratio ]{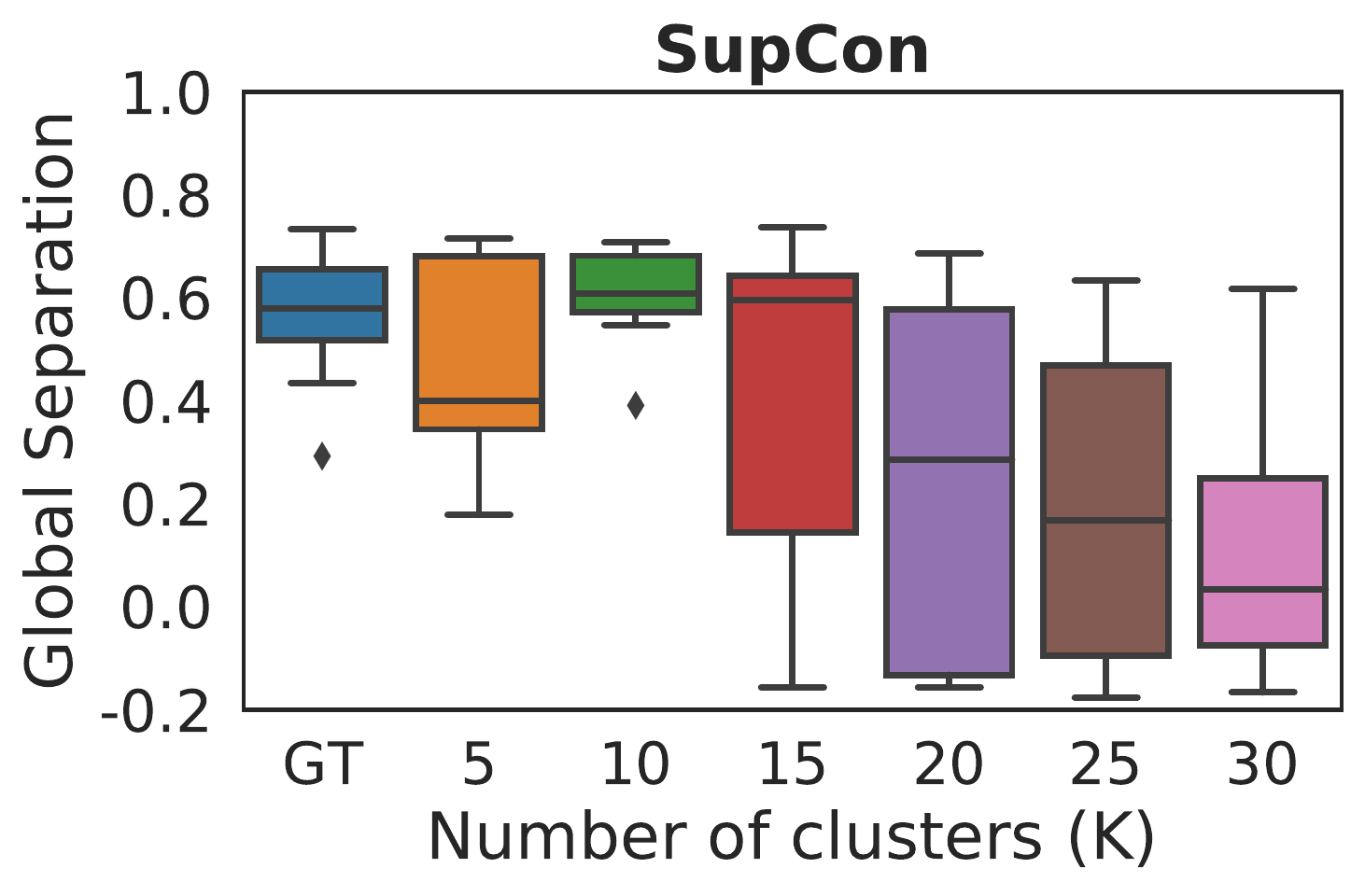}
    \includegraphics[width=0.45\linewidth, keepaspectratio]{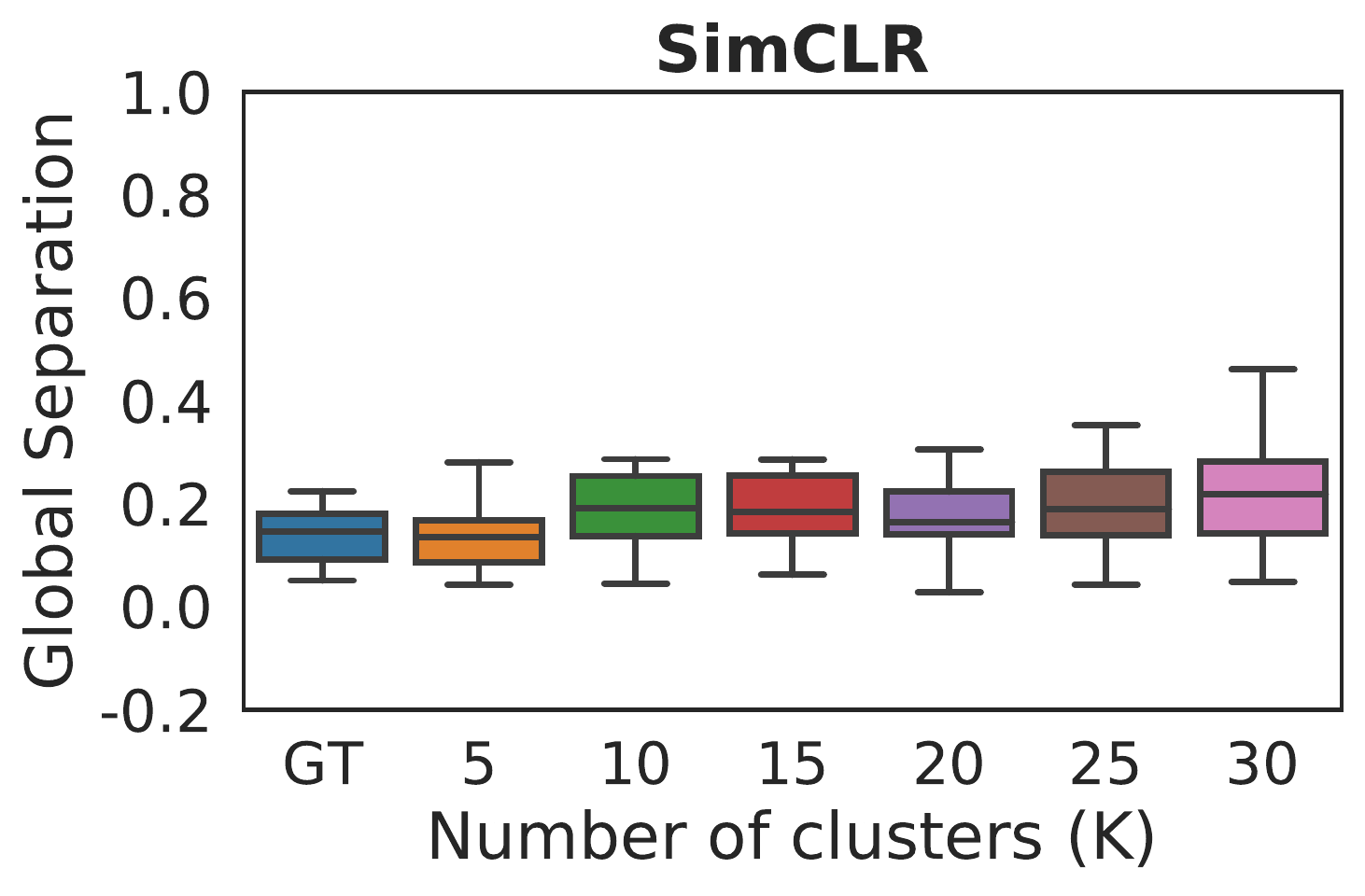}
    \vskip\baselineskip
    \includegraphics[width=0.45\linewidth, keepaspectratio]{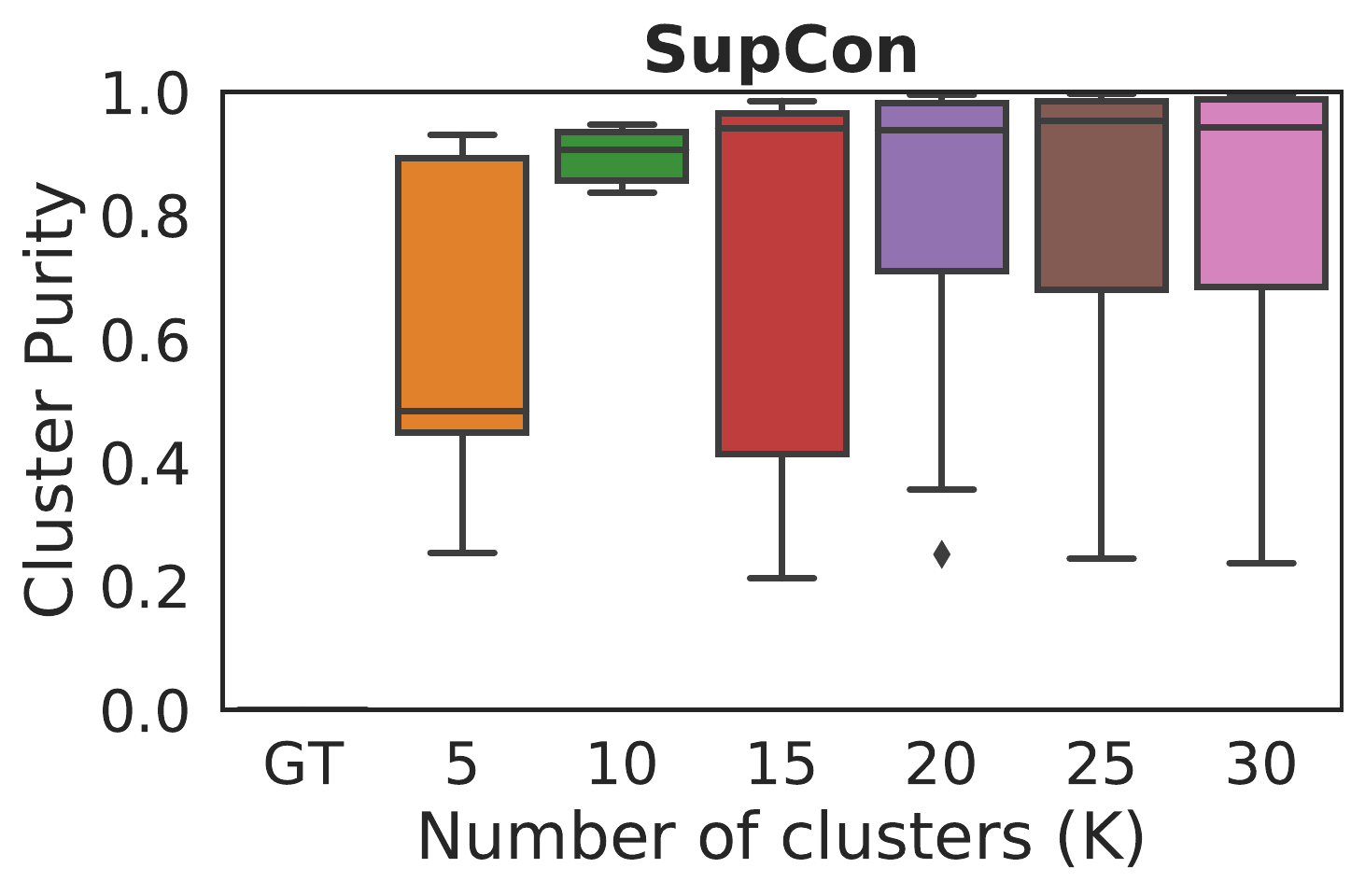}
    \includegraphics[width=0.45\linewidth, keepaspectratio]{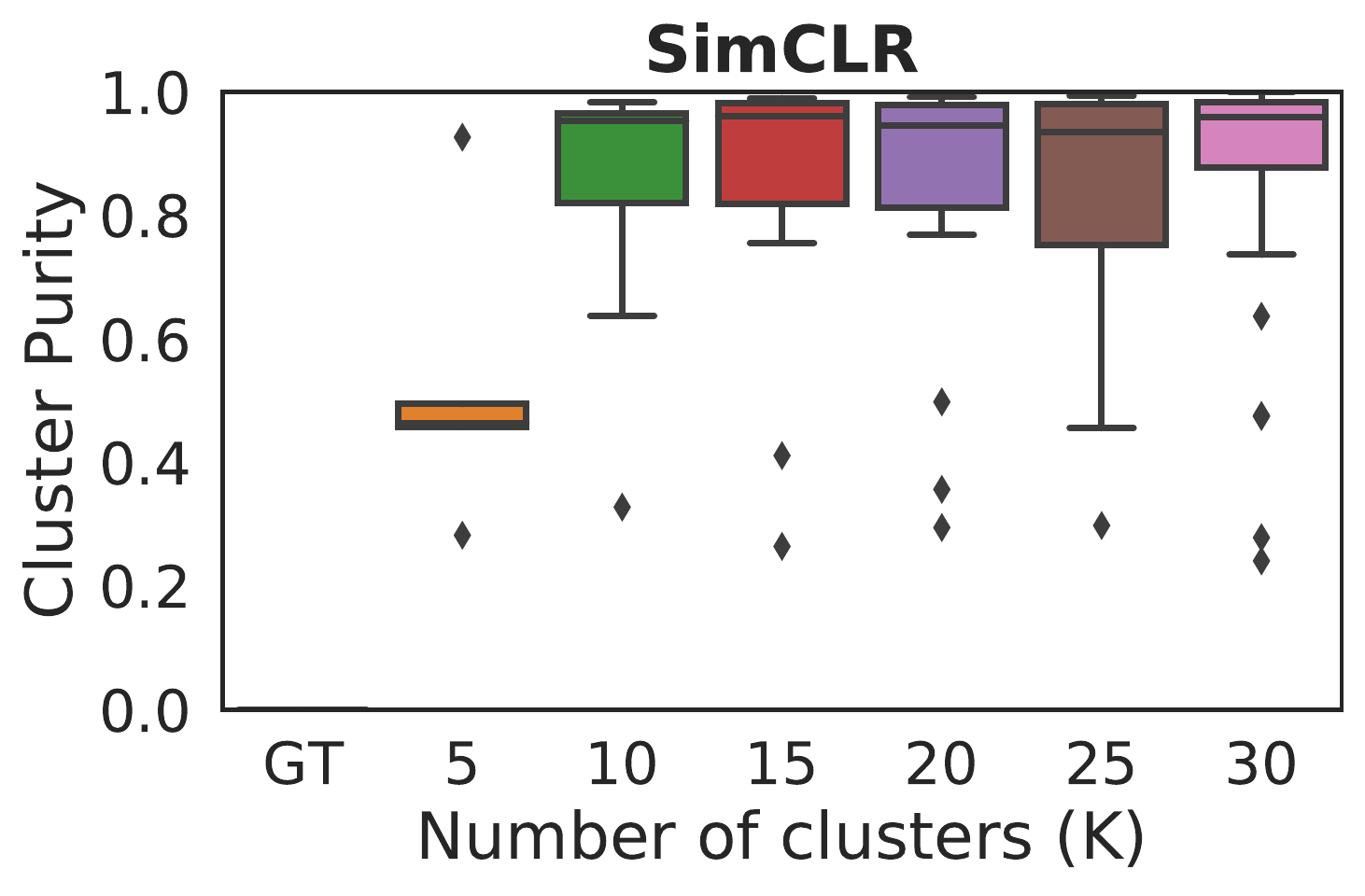}

    \caption{Comparison of cluster quality for SupCon and SimCLR based on GS and CP. Starting with default class-based (GT) clusters, k-means with an increasing number of  clusters  is performed to determine the  cluster quality of underlying feature-based clusters.}
    \label{fig:gs_cp}
\end{figure}

\paragraph{Class-based or feature-based clusters?}
In the previous section, we found no prominent GT \textit{class-based cluster} separation for the unsupervised CL approach. However, it is still possible that \textit{feature-based clusters} form based on common features than class semantics. Thus, we further investigated whether applying k-means clustering on top of such contrastively trained embeddings can regroup them into better separated distinct clusters using the GS and CP metrics along with cosine similarity, as shown in Figure \ref{fig:gs_cp}. Class-based clusters $K = GT$ are shown for comparison. The boxplots indicate the range of global separation for all clusters for a given configuration. CP is only applied to the k-means clusters since the class-based clusters always have a purity of 1.

\begin{figure}[h!]
    \centering
    \includegraphics[width=0.9\linewidth, keepaspectratio]{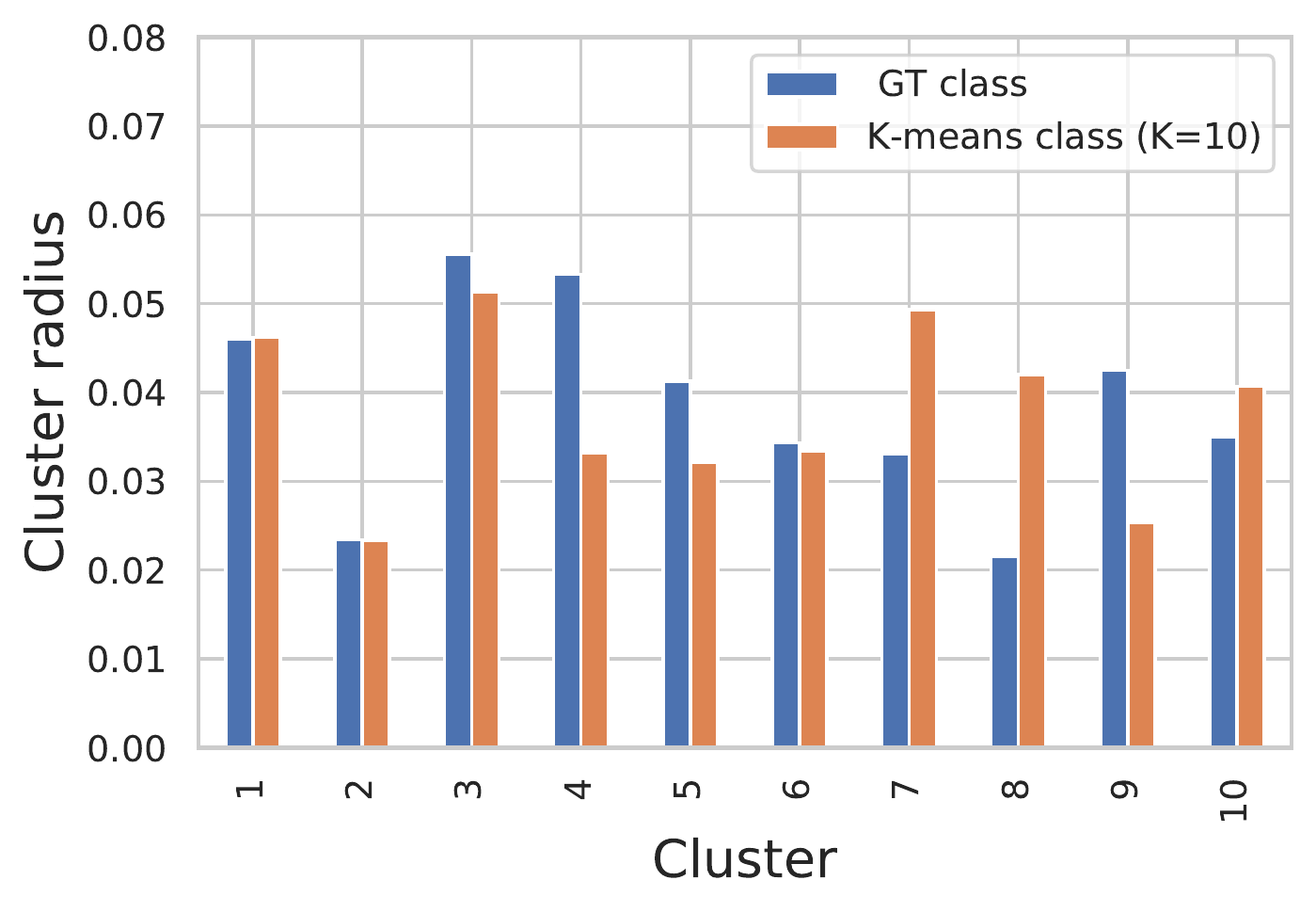}
    \caption{Variation in sizes of clusters in embedding space }
    \label{fig:cluster_radius}
\end{figure}
 
 SupCon shows already good cluster separation for GT class-based clusters due to supervised contrastive training. Subsequently, with 10 k-means clusters, separation gets even slightly better while cluster purity around 90\% indicates a significant overlap with the GT clusters. But, for SimCLR we observe GT class-based clusters are not well separated.  
 This is expected since the contrastive loss is applied at the projection layer maximising its effect there, it is difficult for feature layer to show well separated GT class-based clusters for an unsupervisedly learnt embeddings. 
 
\paragraph{What do clusters look like?}
Many works on OOD detection using the distance-based scores use a single global threshold to distinguish between ID and OOD samples \cite{hsuGeneralizedODINDetecting2020a, leeSimpleUnifiedFramework2018, tackCSINoveltyDetection2020, sehwagSSDUnifiedFramework2021, hendrycksUsingSelfSupervisedLearning2019}, i.e. if a sample is farther away from the closest cluster than this threshold, it is classified as OOD. 
However, this implies the assumption that all clusters have an equal size or even hyperspheres with the same diameter depending on distance metrics used. Since this is a very strong assumption, we compare the size of the individual clusters. Figure \ref{fig:cluster_radius} shows the cosine distance from the cluster centre encompassing 95\% of the training samples of this cluster using the feature layer embeddings from SupCon as an example. We interpret this figure as an approximation of the cluster radius and observe that the largest cluster has almost 2.5 times the smallest cluster. This is true for GT clusters as well as k-means clusters. We therefore conclude that the use of a single global threshold does not reflect the true nature of the clusters motivating the use of individual thresholds for each cluster. Thus, in the next section, we present a study on the OOD detection performance for different models and metrics by evaluating with both per cluster-based as well as global threshold based metrics. 

\section{OOD detection based on clustering}
\label{sec:ood}
Here, we describe our OOD detection method using cluster-based thresholds and compare them to global thresholds for different distance-based scoring metrics and clustering methods across diverse ID/ OOD datasets and model architecture. Further, we investigate the relationship between cluster quality and OOD detection performance. 
 
 \subsection{Method} 
 \label{sec:ood_method}
 From our analysis in Section \ref{sec:cluster}, we can find well-separated clusters of ID samples at the last layer of the feature extractor once a model is sufficiently trained. We use the distance to the mean of a cluster as an indication of how similar a test sample is to one of the samples in ID cluster. The farther the sample, the less likely it is to be related to ID clusters. If a sample is far enough away from all clusters, it is considered to be an OOD sample. Since we discovered in the previous section that clusters have varying sizes, we propose to employ \textit{cluster-based thresholds} as compared to global thresholds for  distance-based scoring metrics.
 
 Our \textit{approach} consists of the following steps: (1) During training, the mean of all clusters with respective training samples is calculated. This is for clusters based on GT class labels. In case of k-means clustering or using Guassian Mixture Models (GMM), mean is calculated based on samples assigned to respective clusters by these methods. (2) Using mean and distance metrics, distance scores are calculated for all train and test samples. (3) During inference, for each cluster, the set of distance scores for the respective reference distribution (train/test) are taken. For each new test sample, a probability score is assigned depending upon where the distance score of the given test sample can fit in the overall distribution of distance scores of the given cluster. These probability scores of test samples are finally used for calculating evaluation metrics.  
 The global threshold based probability scores can also be similarly calculated by taking the entire reference distribution of distance scores into account.

 
 \paragraph{Distance metrics:}
 In order to test the proximity of a sample to a cluster, we need to either approximate the underlying distribution function or use a simple distance metric such as \textit{Cosine similarity} or \textit{Euclidean} distance. The former can be analysed by mapping to cluster conditioned linear multivariate Gaussians. This can be expressed as \textit{Mahalanobis distance} based score \cite{mahalanobis1936generalized}, by calculating cluster-wise mean ($\mu_c$)  and co-variance ($\Sigma_c$) corresponding to the features $f(x)$ of a test sample $x$,  as given in Equation \ref{eq:mahalanobis}. Here $S_c(x)$, represents the distance of a sample $x$ from centre of cluster $c$. 
 \begin{equation}\label{eq:mahalanobis}
     S_c(x) = (f(x) - \mu_{c})^{T}  \Sigma^{-1}_{c} (f(x) - \mu_{c}) 
 \end{equation} 
 Although Mahalanobis distance is a reasonable choice for highly correlated data it is prone to the \textit{curse of dimensionality} with increasing dimensions of the embeddings \cite{4815271} as well as increasing components.
 Cosine similarity, is often deemed to be better suited for computing distances with high dimensional inputs. Since, during contrastive training, instance based comparisons utilise cosine similarity, we employ this score for cluster based distances as well.
 In our experiments, we have also used GMM to map the embedding space into a mixture potential Gaussian components, thus assigning each sample to one of these clusters/components based on Expectation Maximisation rather than simply taking the GT class labels. Subsequently, Mahalanobis distance has been used to calculate the distance scores with respect to this cluster assignment. We have incorporated this an alternate to k-means clustering which simply utilises Euclidean distance between the feature vectors to assign samples to different clusters. Finally,  we compare all the cluster-based and global evaluation metrics with a global evaluation of the probability scores assigned by applying GMM irrespective of which clusters/components each sample belong to.

  \begin{figure}[h!]
    \centering
    \includegraphics[width=\linewidth, keepaspectratio]{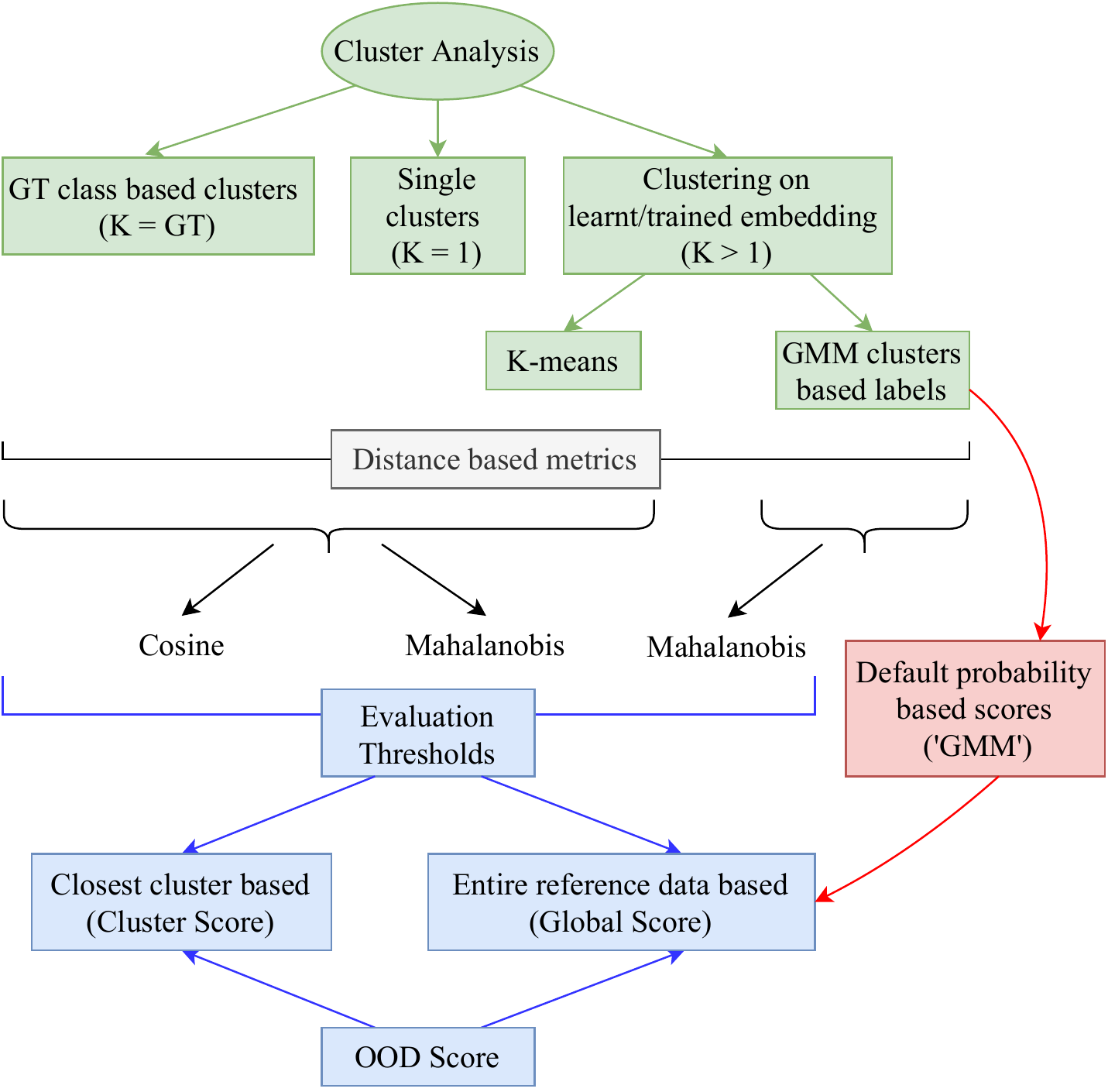}
    \caption{Schematic showing our OOD evaluation pipeline }
    \label{fig:ood_eval_pipeline}
\end{figure}

 \subsection{Experiments and Discussion}
 \label{sec:ood_exp}
 
\begin{figure*}[hbtp]
 \centering
    \includegraphics[width=0.45\textwidth, height=3.5cm ]{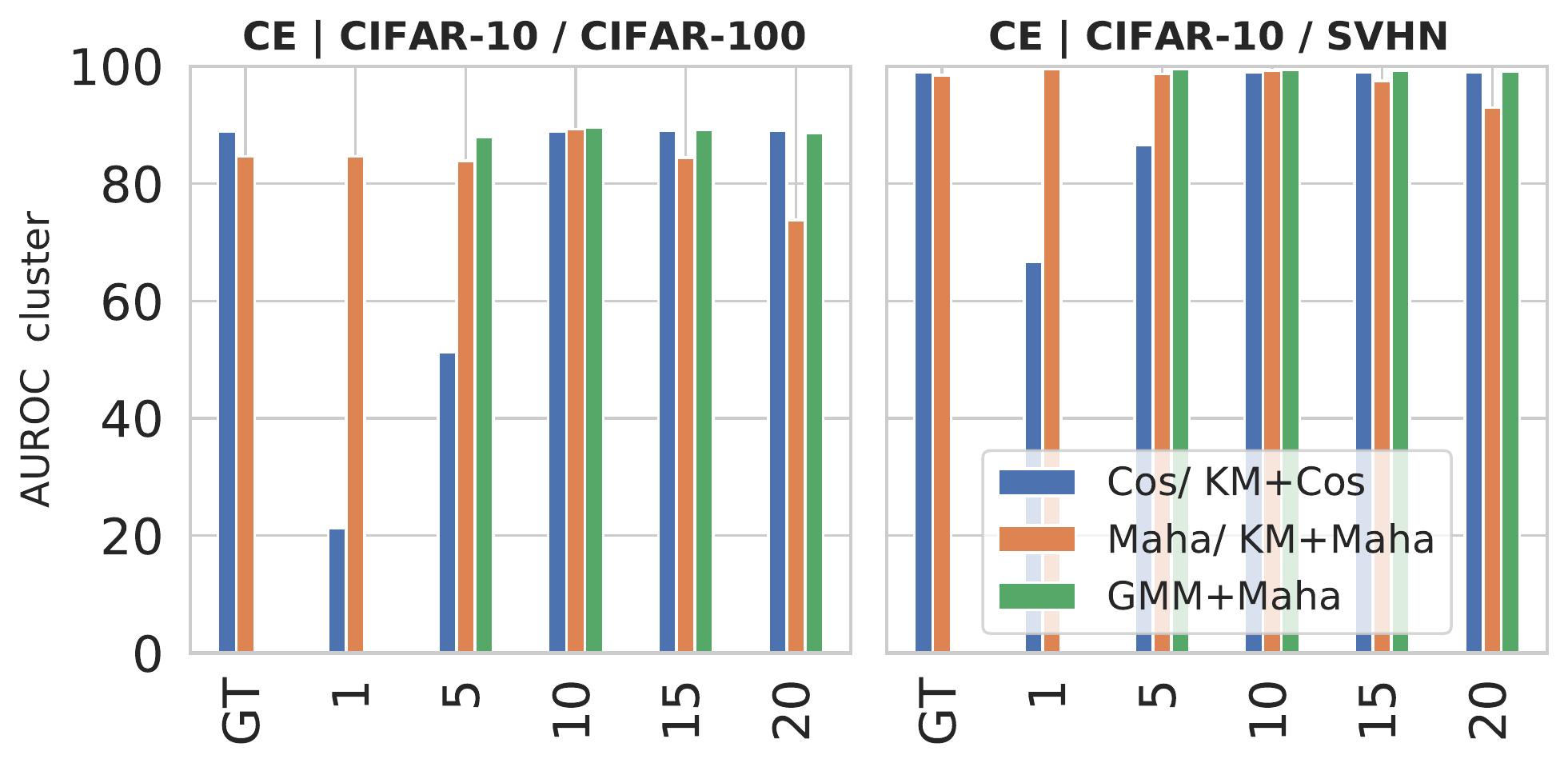}
    \includegraphics[width=0.45\textwidth, height=3.5cm ]{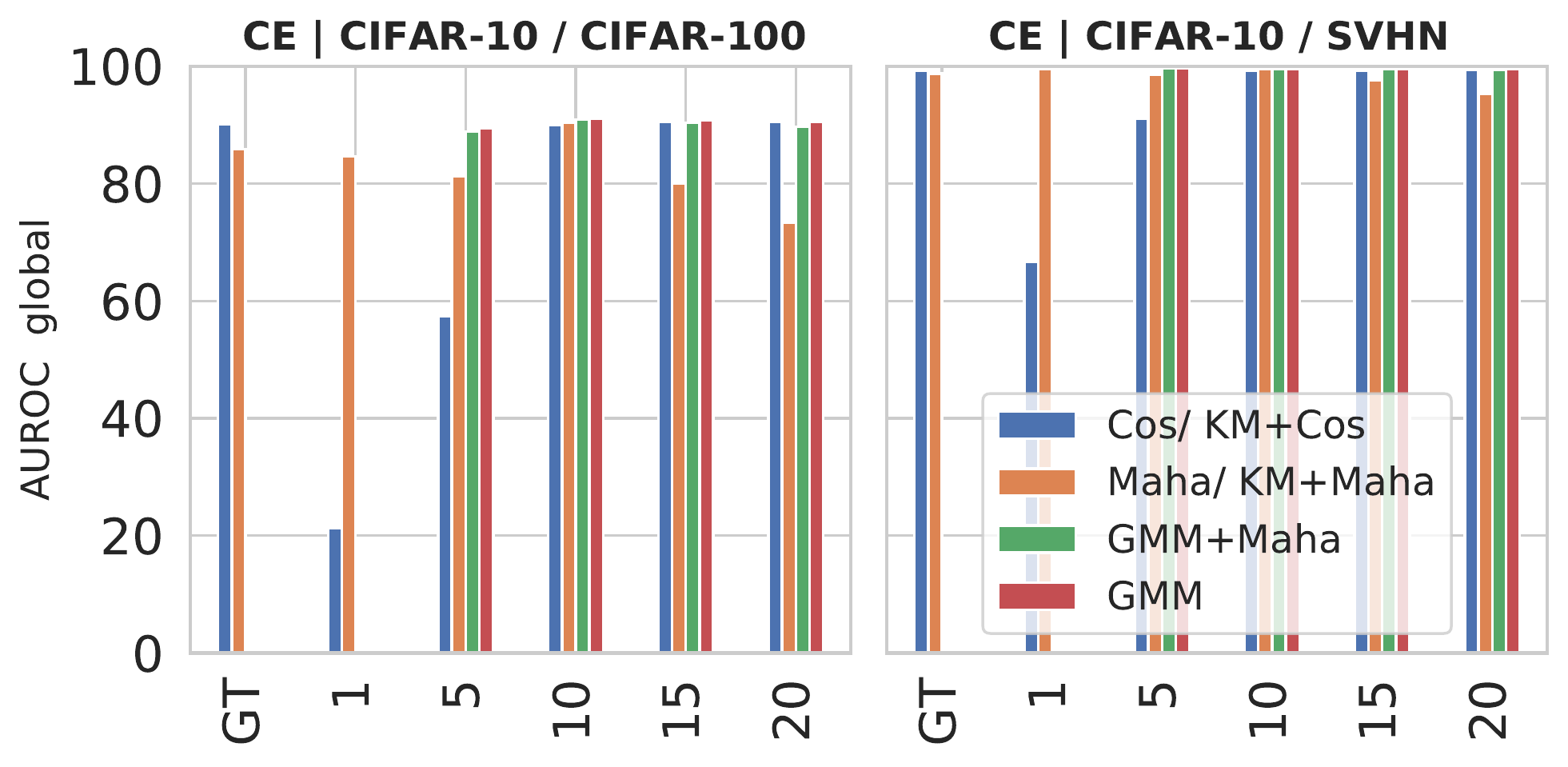}
    \includegraphics[width=0.45\textwidth, height=3.5cm ]{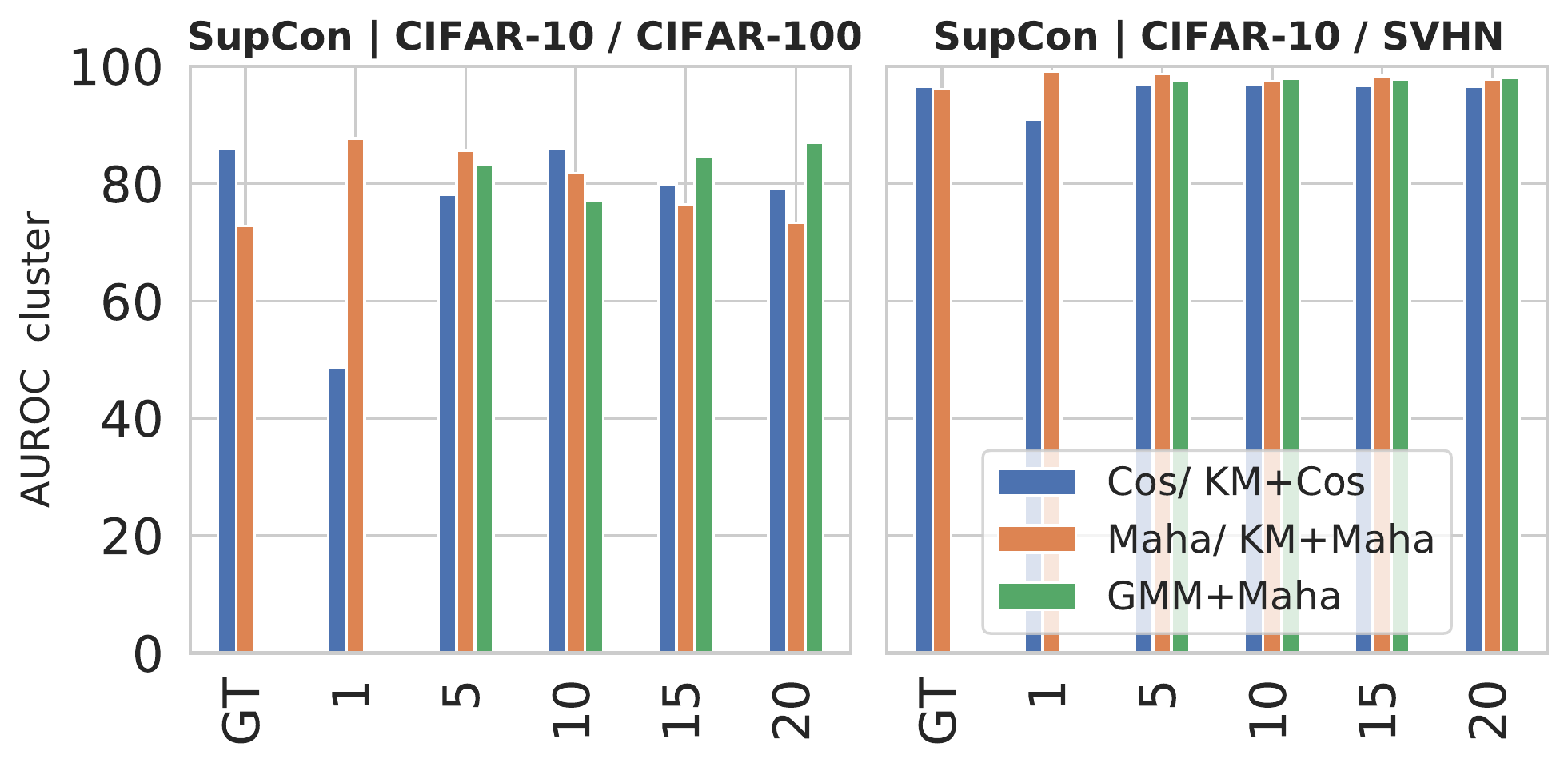}
    \includegraphics[width=0.45\textwidth, height=3.5cm ]{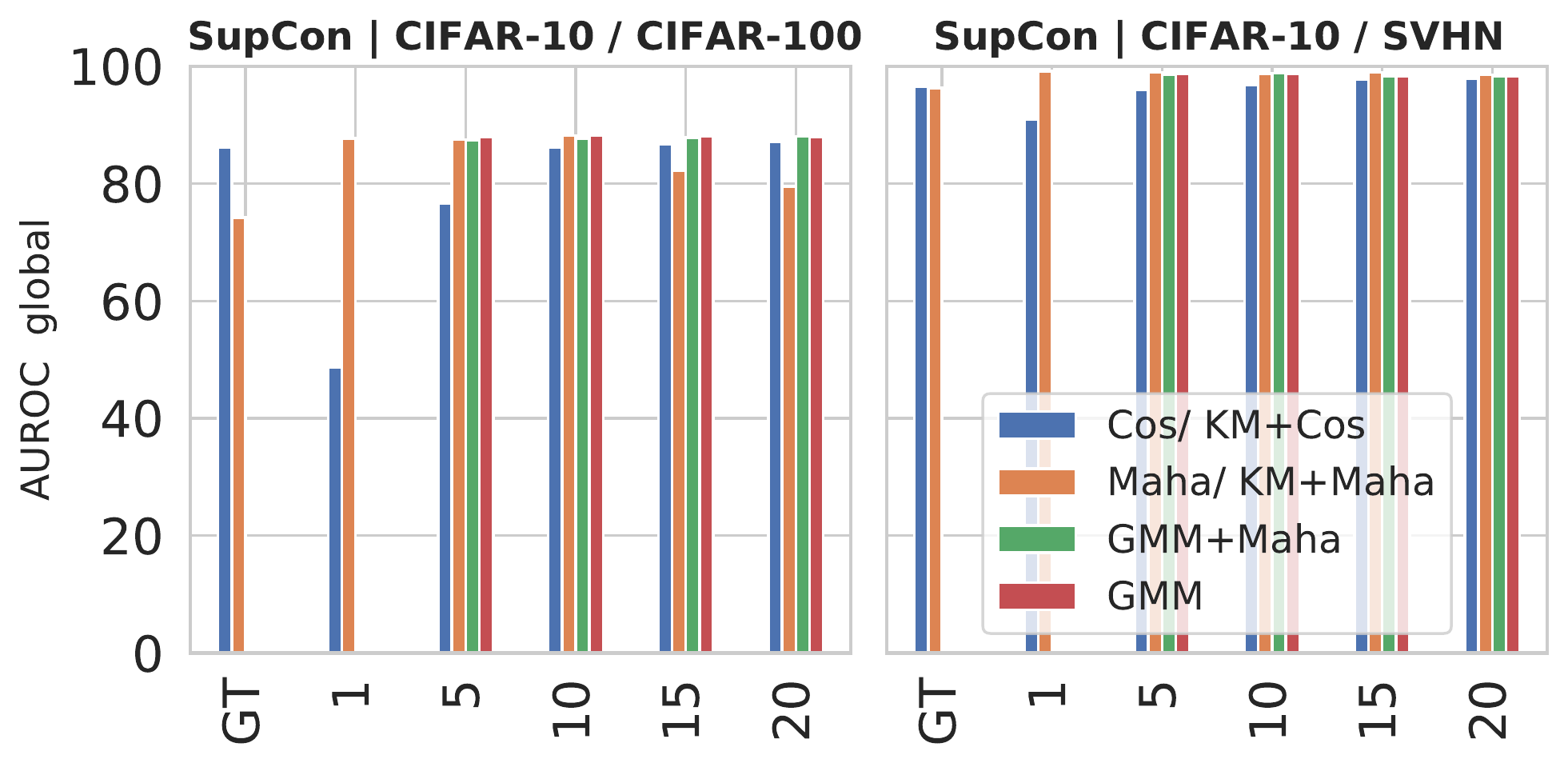}
    \includegraphics[width=0.45\textwidth, height=3.5cm ]{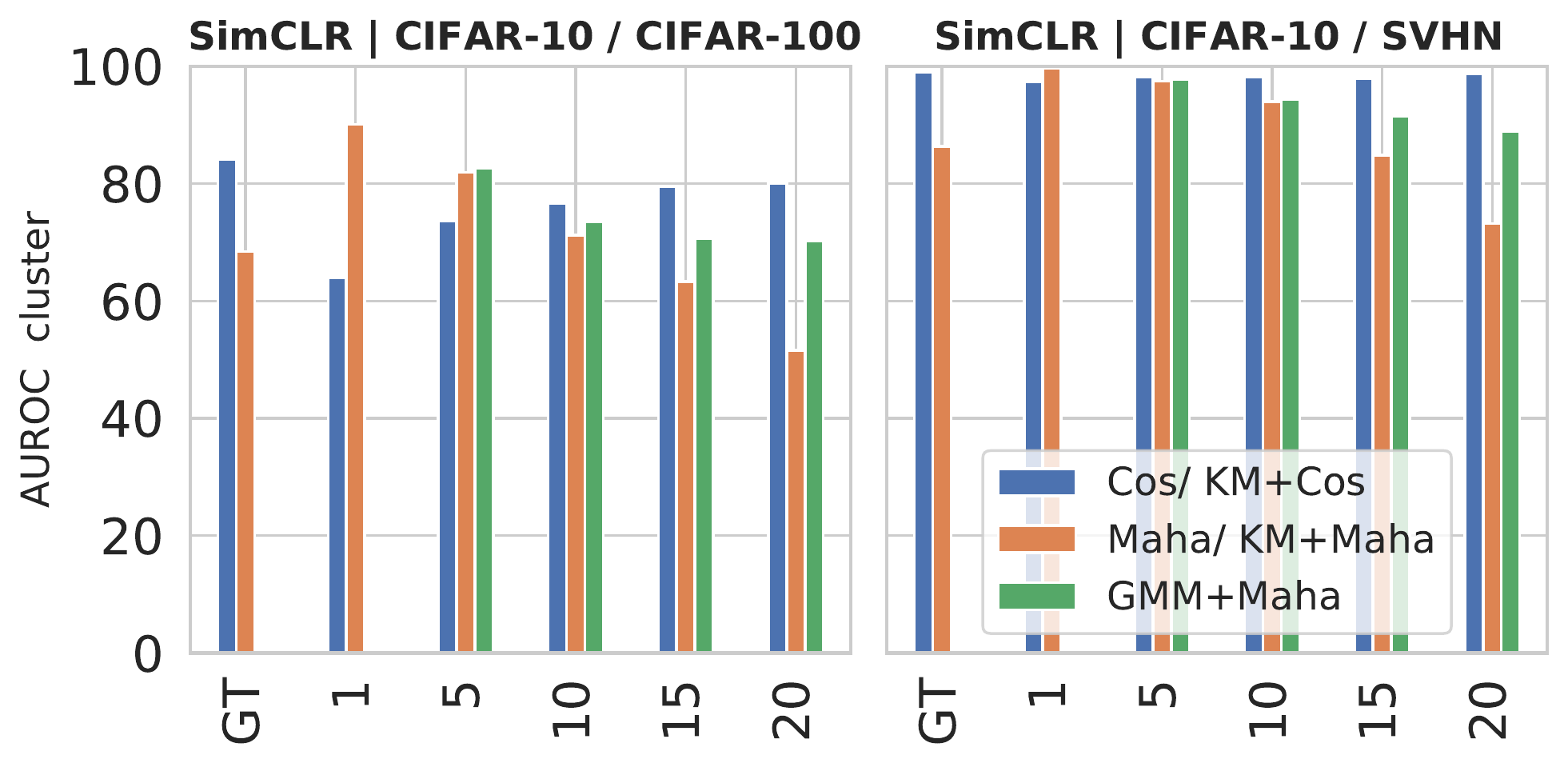}
    \includegraphics[width=0.45\textwidth, height=3.5cm ]{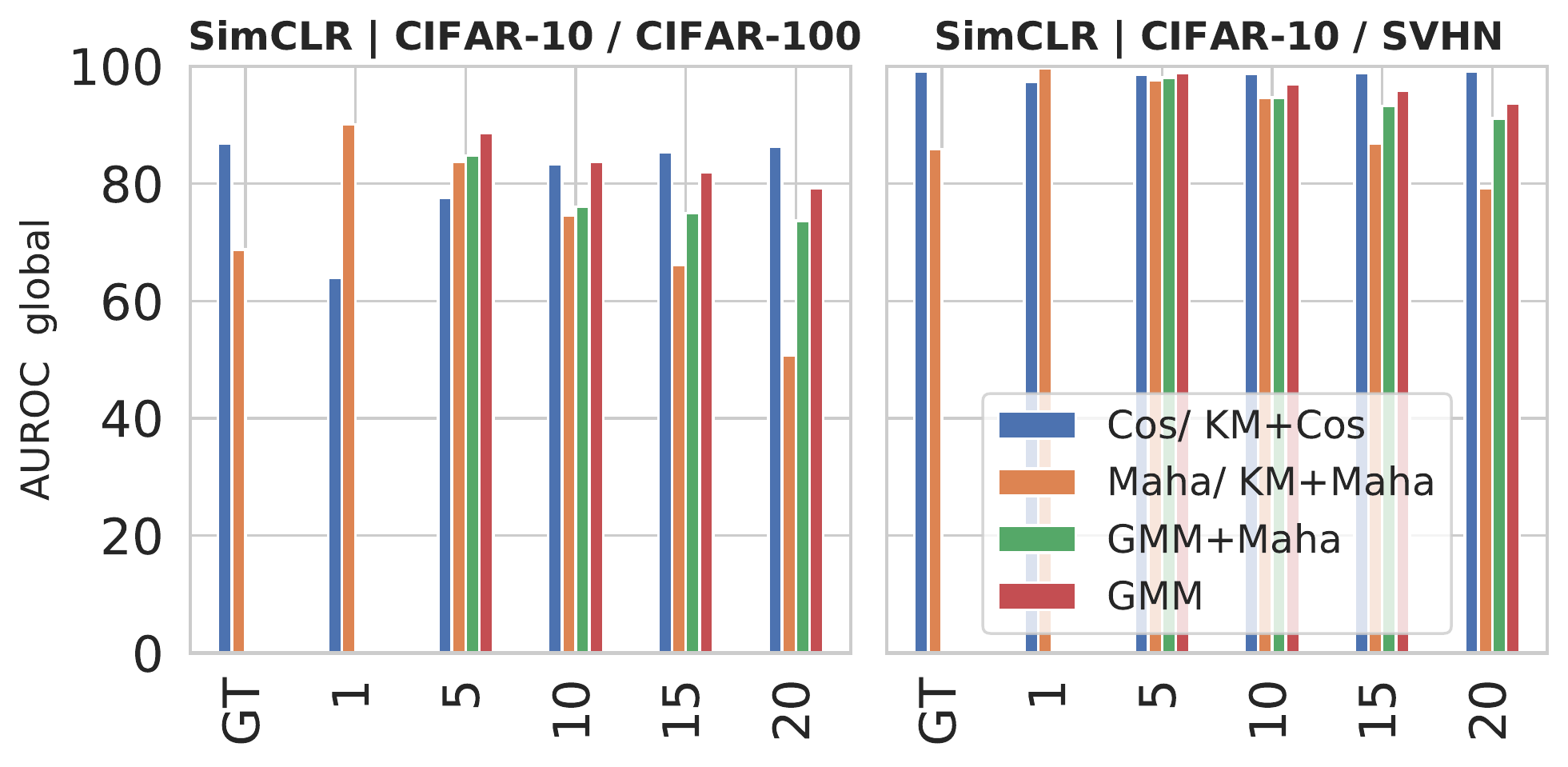}  
    \includegraphics[width=0.45\textwidth, height=3.5cm ]{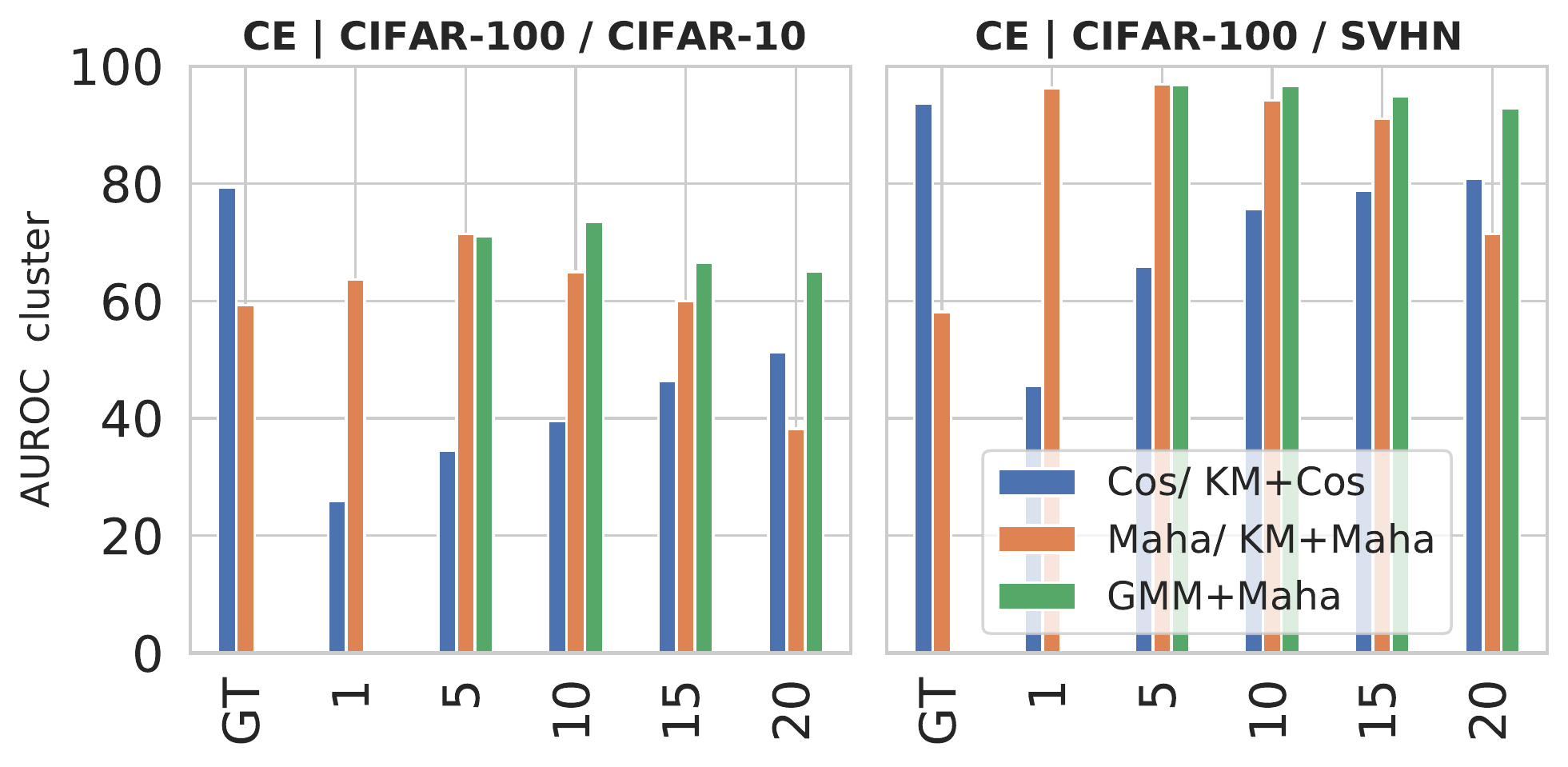}
    \includegraphics[width=0.45\textwidth, height=3.5cm ]{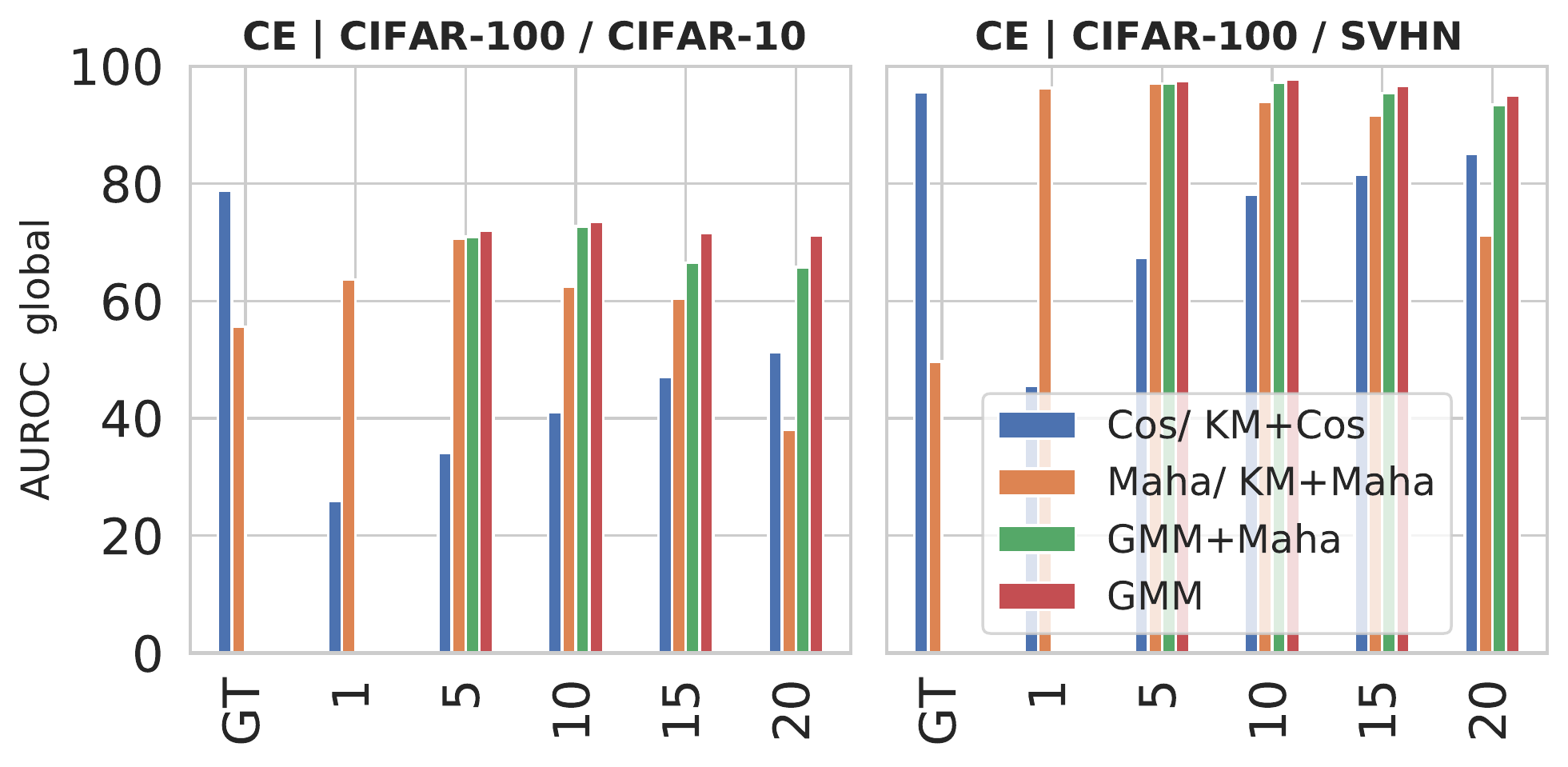}
    \includegraphics[width=0.45\textwidth, height=3.5cm ]{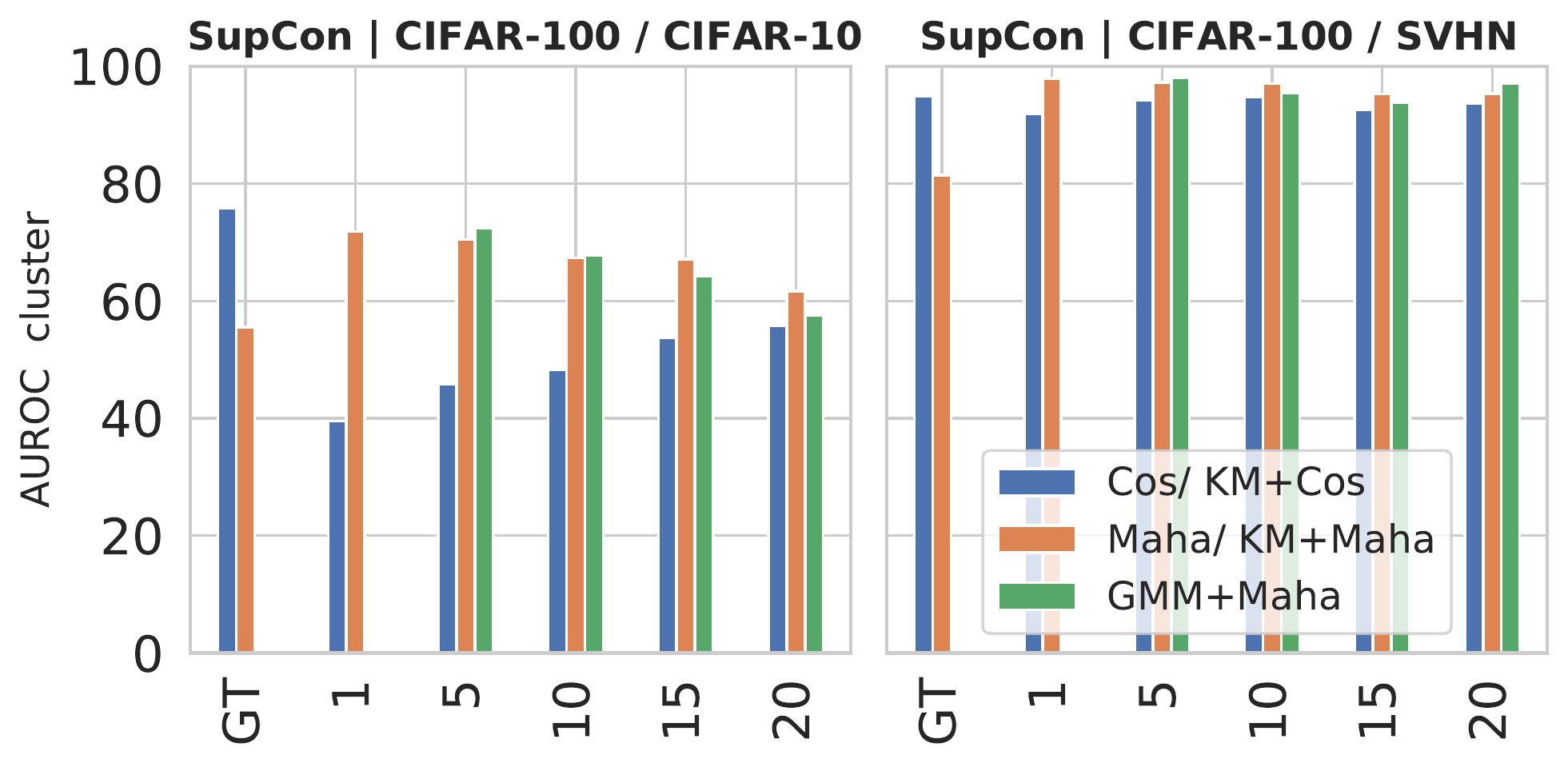}
    \includegraphics[width=0.45\textwidth, height=3.5cm ]{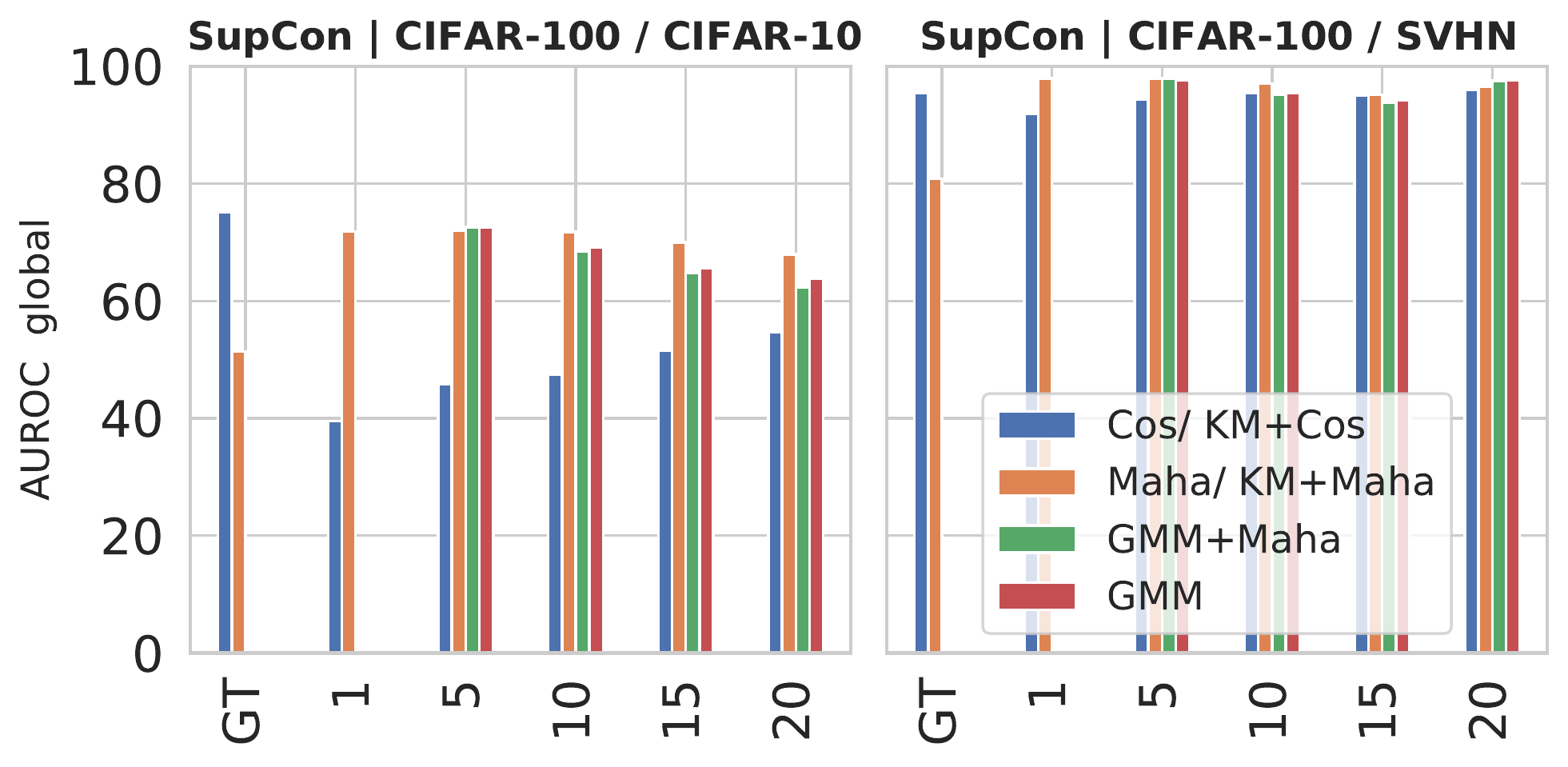}
    \includegraphics[width=0.45\textwidth, height=3.5cm ]{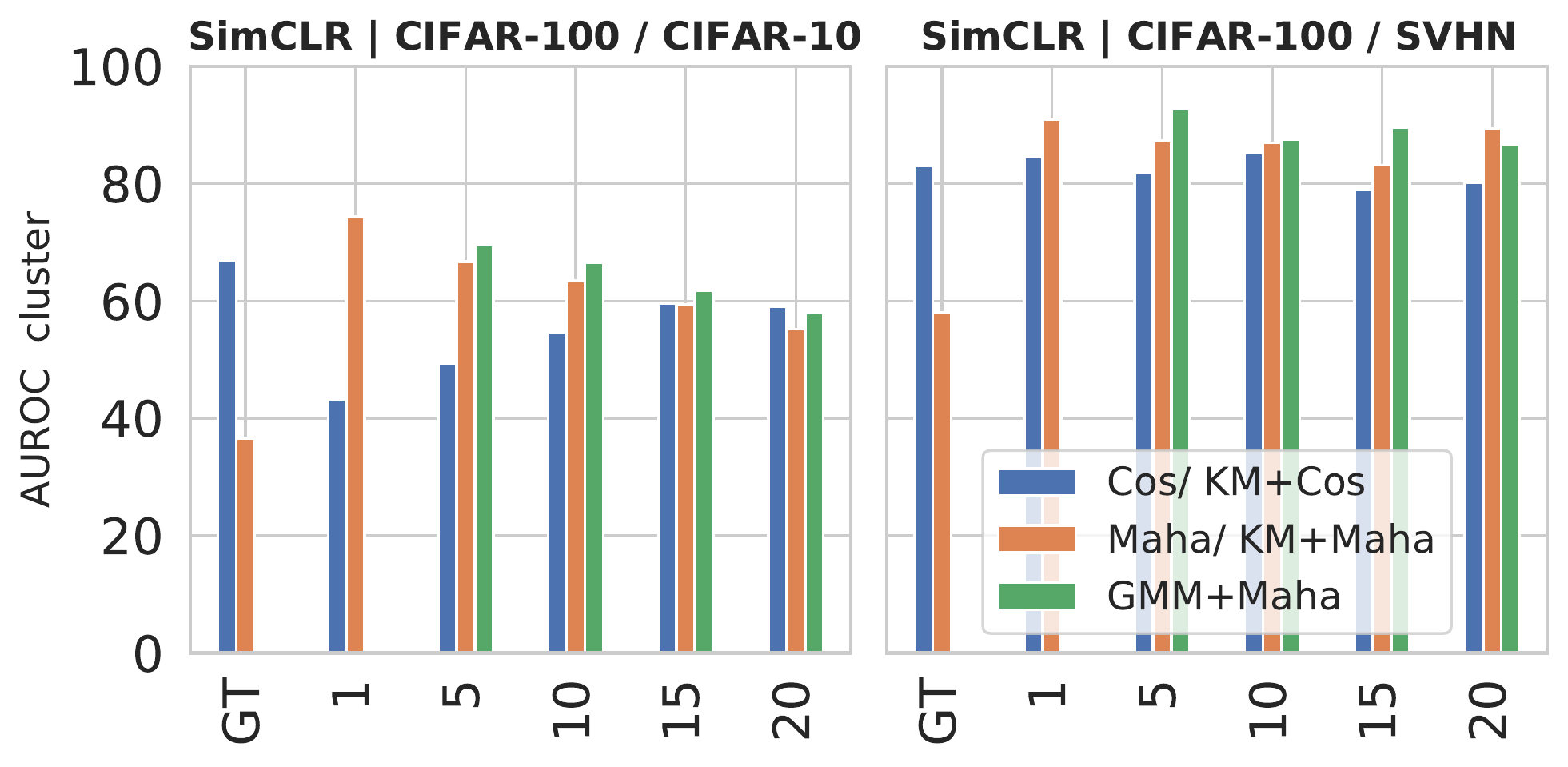}
    \includegraphics[width=0.45\textwidth, height=3.5cm ]{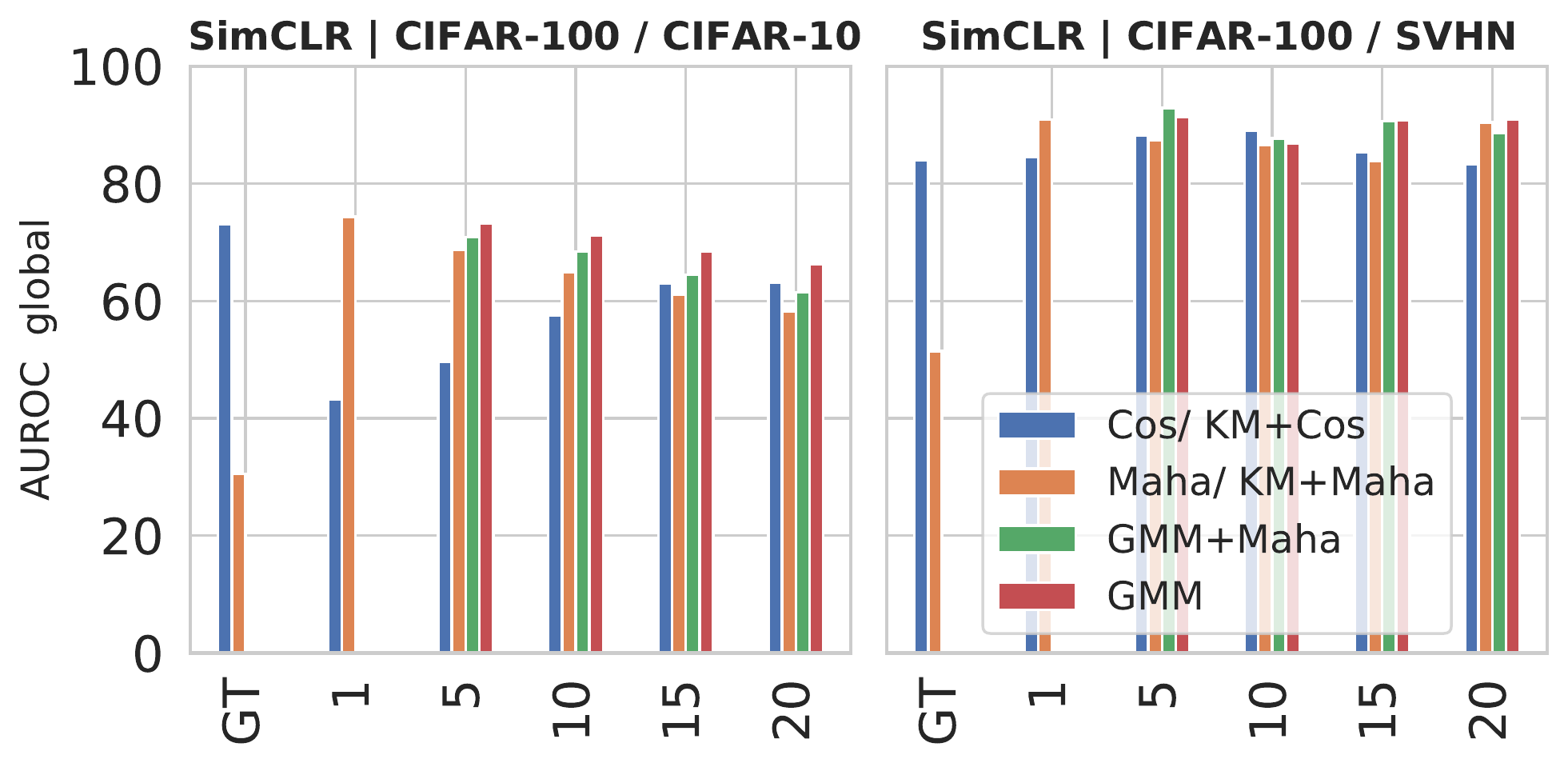}  
    
    \caption{Comparison of OOD detection performance using cluster-based and global thresholds for AUROC evaluation metric on CIFAR-10 (top 3 rows), CIFAR-100 (bottom 3 rows) as ID data across different methods(CE, supCon, SimCLR)  for different distance metrics. $x$ axis shows number of clusters (K), where GT implies GT number of clusters, $ K=1 $ implies single cluster and $K>1$ indicates number of clusters after k-means/ GMM clustering. }
    \label{fig:auroc_bar_resnet50}
\end{figure*}

  In this section, we investigate the OOD performance for models trained on CIFAR-10, CIFAR-100 \cite{krizhevsky2009learning} with SupCon, SimCLR and baseline CE using ResNet architecture. We present results with SVHN \cite{netzer2011reading} and CIFAR-100 /CIFAR-10 as OOD data sets due to limited space, however we have conducted experiments with other OOD datasets like resized ImageNet, LSUN \cite{liangEnhancingReliabilityOutofdistribution2020} and our observations are also extended to these datasets. The results of our study is presented in Figure \ref{fig:auroc_bar_resnet50}.
  
  Given 'K' represents the number of clusters,  we base our investigation on following type of clusters in the embedding space - GT class based clusters ($K=GT$) where features are taken from embeddings trained based on GT class labels, single cluster ($K=1$) where all the embeddings are taken as one cluster and finally similar feature-based clusters  ($K = 5, 10, 15, ..$) mapped by using either k-means clustering or GMM to re-group the embeddings into distinctive clusters. This process of the OOD evalaution pipeline has been represented in the schematic given in Figure \ref{fig:ood_eval_pipeline}. While $K=1$ case is similar to the results presented in \cite{sehwagSSDUnifiedFramework2021} for SimCLR, however by investigating further clusters we show that the mentioned setting is not always the best case scenario. It largely depends on choice of different variables as shown in our study. As mentioned in previous Section \ref{sec:ood_method}, for all cluster based analysis using k-means, we employ Cosine similarity (\lq KM+Cos\rq)and Mahalanobis distance (\lq KM+Maha\rq) except for clusters based on GMM based target labels ( \lq GMM+Maha \rq) we use only Mahalanobis distance only due to inherent assumption of Gaussian components. Finally, we employ the Area under ROC curve (AUROC) as the main evaluation metric for OOD detection where we present \lq AUROC cluster \rq, 
  \lq AUROC global \rq as the respective scores for cluster-based and global thresholds.
  We also compare with default \lq GMM\rq scores in global evaluation.

\paragraph{Performance across GT class-based and feature-based clusters:}
From Figure \ref{fig:auroc_bar_resnet50}, using class-based clusters ($K=GT$), we observe comparable AUROC scores across all the methods with SupCon quite similar to CE and slightly better than SimCLR. For feature clusters with $K=10$, we achieve almost similar performance as with class-based clusters for CIFAR-10. This is expected since we see a strong overlap of the clusters generated by k-means/ GMM and the GT classes. With further increasing clusters ($K>10$), almost similar trend exists as $K=10$. For CIFAR-100 as ID, we see an overall decreasing trend across all distance metrics except for Cosine ($K>1$). This could be due to fewer clusters than GT cluster(100). However, for SimCLR it appears to be beneficial to treat all ID samples as one single cluster ($K=1$), as illustrated in Figure \ref{fig:supcon_simclr}. This result is in line with our observation, that SimCLR does not lead to well-separated clusters and therefore making a distinction between clusters is superfluous. Nonetheless, it is a bit surprising that this single clusters also leads to competitive OOD detection performance across all datasets. 

\paragraph{Performance across distance metrics:}

Taking cue on SimCLR performing best at $K=1$  using Mahalanobis distance in almost all cases in Figure \ref{fig:auroc_bar_resnet50} indicates the possibility of high covariance between clusters when taken as a whole. Notably, for GT clusters cosine similarity always performs better than Mahalanobis \cite{4815271}. However, this is not the case for $K>1$ as Mahalanobis shows a decreasing trend in general from maximum at single cluster (for SimCLR) while the AUROC for cosine similarity drops for $K=1$ and then continues an upward trend until optimum cluster at $K=10$ for SupCon and slightly further for SimCLR in case of CIFAR-10 as ID dataset. For CIFAR-100, it continues with an upward trend for both SupCon and SimCLR. CE with cosine similarity achieves best performance across all methods in CIFAR-100. Also for clustering cases, performance of GMM and k-means cluster based Mahalanobis has been comparably same with slightly greater performance by GMM based clusters. For supervised cases, the AUROCs remain comparable with increase in clusters but for SimCLR they show usual declining trend. For most global cases, the default GMM probability remain steady and similarly high to other distance metrics across different clusters. 

\paragraph{Performance across cluster-based and global thresholds}
Although cluster-based thresholds seek to represent the embedding clusters better than global thresholds, however global AUROCs tend to perform slightly better than cluster AUROCs in most cases. Although the clusters are of inequal sizes, however depending on features of the given ID dataset they tend to be quite overlapping, so that global thresholds seem to be good enough for OOD detection.  

\paragraph{Performance across OOD datasets:}
All the above observations remain consistent across all the OOD datasets. However, we note SVHN being semantically quite different from CIFAR-10 (Far OOD) achieves much better AUROC as compared to CIFAR-100 which is semantically quite close to CIFAR-10. 

\paragraph{Performance across ID datasets}
We note that the AUROCs for CIFAR-100 (ID) vs CIFAR-10 (OOD) are much lower compared to vice-versa, although it follows similar trends mentioned above. This could be due to more classes (100 vs 10) leading to much smaller closely overlapping embedding clusters. This distinction becomes more difficult with similar OOD dataset like CIFAR-10, as when compared to really different SVHN.

\paragraph{Performance across model architectures}
We conducted similar experiments on ResNet-18 (although not reported here) vs ResNet-50, however we find much lesser overall AUROCs in the former, with SimCLR performing superior than supervised cases. This could potentially indicate requirement for bigger models as supervised cases require higher positives for instance based discrimination in CL.


\section{Conclusion}
In this work, we first investigated the nature of clusters in embedding spaces of contrastively trained models for image classification. We found that supervised contrastive training leads to well-separated clusters of in-distribution data in the embedding space and that these clusters correlate strongly with the ground truth classes. Unsupervised contrastive training on the other hand leads to mostly overlapping clusters that cannot be clearly distinguished. To our surprise, standard cross-entropy loss also lead to reasonably distinct clusters. Secondly, we proposed a modular OOD detection method exploiting proximity of similar samples in the embedding space allowing us to compare different distance metrics, clustering methods and thresholding strategies  across a selection of model architectures, ID and OOD datasets. While we could reproduce the superior performance of SimCLR with a single cluster and Mahalanobis distance for CIFAR-10 vs. CIFAR-100, cross-entropy with clusters based on ground truth classes and cosine similarity performed best when the in- and out-of-distribution roles are reversed. However, in many cases there exist several combinations that lead to similar detection performance making the results even more ambiguous. We therefore have to conclude, that there is no clear winner and not even a solid trend, yet. For a deeper understanding, future work should extend this investigation to a broader and more diverse set of model architectures. Given recent observations that many OOD methods do not translate well from academic datasets to real world applications~\cite{berger2021confidence}, further research should focus on studies with real image data from medical, automotive or industrial use cases. 
\bibliography{aaai22}
\section{Acknowledgments}
This work was funded by the Bavarian Ministry for Economic Affairs, Regional Development and Energy as part of
a project to support the thematic development of the Institute for Cognitive Systems (IKS).

\end{document}